\documentclass{article} % For LaTeX2e
\usepackage{iclr2015,times}
\usepackage{hyperref}
\usepackage{url}
\usepackage{amssymb}
\usepackage{amsthm}
\usepackage{verbatim}
\usepackage{graphicx}
\usepackage{amsmath}
\usepackage{subcaption}

\title{Neural Network Regularization via Robust Weight Factorization}

\author{
Jan Rudy, Weiguang (Gavin) Ding, Daniel Jiwoong Im \& Graham W.~Taylor\\
School of Engineering\\
University of Guelph\\
Guelph, Ontario, Canada \\
\texttt{\{jrudy,imj,wding,gwtaylor\}@uoguelph.ca} 
}

\newcommand{\x}{\mathbf x}
\newcommand{\y}{\mathbf y}
\newcommand{\z}{\mathbf z}
\newcommand{\W}{\mathbf W}
\newcommand{\U}{\mathbf U}
\newcommand{\V}{\mathbf V}
\newcommand{\h}{\mathbf h}
\newcommand{\e}{\mathbf e}
\newcommand{\rand}{\mathbf r}

\newcommand{\bias}{\mathbf b}
\newcommand{\Reals}{\mathbb{R}}

\iclrconference % Uncomment if submitted as conference paper instead of workshop

\begin{document}

\maketitle

\begin{abstract}
  Regularization is essential when training large neural networks.  As deep neural networks can be mathematically
  interpreted as universal function approximators, they are effective at memorizing sampling
  noise in the training data. This results in poor generalization to unseen data.
  Therefore, it is no surprise that a new regularization technique, Dropout, was 
  partially responsible for the now-ubiquitous winning entry
  to ImageNet 2012 by the University of Toronto.
  Currently, Dropout (and related methods such as DropConnect) are
  the most effective means of regularizing large neural networks. These
  amount to efficiently visiting a large number of related models at training time, while
  aggregating them to a single predictor at test time. The proposed
  FaMe model aims to apply a similar strategy, yet learns a factorization
  of each weight matrix such that the factors are robust to noise.
\end{abstract}

\section{Introduction}
Much of the recent surge in popularity in neural networks, especially in 
their application to classification of visual data, is due to advances 
in regularization. The winning entry in the 2012 ImageNet LSVRC-2012 challenge
by \citet{krizhevsky2012imagenet} used a deep convolutional
neural network to surpass the competition by a margin of nearly 8\% in top-5
test error rate. They partially attribute their performance to regularization
technique called ``Dropout'' \citep{krizhevsky2012imagenet,hinton2012improving,srivastava2014dropout}.
As large neural networks are extremely powerful, avoiding overfitting is crucial
to increasing generalization performance. Dropout is an elegant and simple solution which is
equivalent to training an exponential number of models. At test time,
these models are `averaged' into a single `mean' predictor which
generalizes better to unseen test data.

However, models with rectified linear activations (ReLU) are known to lead
to sparse activations, with 50\% percent of units with true zero activation \citep{glorot2011deep}. Our experiments have shown that this number is as high as 75\% when training with Dropout.
Although sparse representations are desirable in general \citep{lee2008sparse, ranzato2006efficient, ranzato2007sparse}, they reduce the effective number and size of models that Dropout 
visits during training. In other words, the multiplicative noise applied
by Dropout has no effect on the sparse activations. \citet{ba2014making} address
this by adding an post-activation bias to the ReLU activation function,
making explicit the distinction between sparse units and units masked
by Dropout.

We propose a related regularization procedure which we call \textbf{Fa}ctored \textbf{Me}an
training (FaMe). From a high-level, FaMe is
similar to Dropout. Both methods address the problem of overfitting by efficiently
training an ensemble of models which are averaged together at test time.
Where Dropout achieves this by randomly masking units, FaMe does so by
learning a factorization of each weight matrix such that the factors
are robust to noise. This leads to a more accurate model averaging procedure
without sacrificing FaMe's ability to make use of shared information between the models
visited at training time.

\section{Background}

In this section, we briefly review traditional methods to improve
generalization before discussing modern advances such as Dropout and
Drop-connect. We also introduce the concept of gating: networks
which achieve weight modulation via multiplicative interactions among variables.

\subsection{Traditional approaches to improve generalization}
Traditional approaches to improving generalization in neural networks can be seen as a means of limiting
the capacity of the model.
These include early stopping, weight decay ($L_1$ or $L_2$), weight constraints
or the addition of noise to the during training.
Where weight decay involves adding a proportional to the $L_1$ or $L_2$ norm
of the weights to the objective, weight constraints limits the $L_2$ norm
for the
incoming weight vector of each unit \citep{hinton2012improving}.

Addition of noise
during training is also an effective regularizer \citep{bishop1995training,vincent2008extracting,hinton1993keeping}.
Adding small amounts of noise to the input vector has been shown to be equivalent to Tikhonov regularization
\citep{bishop1995training}. 
%A related approach has been applied to autoencoders.
Denoising autoencoders (DAE) \citep{vincent2008extracting} apply either
additive or multiplicative (e.g.~masking) noise to the input
signal.
The model then must learn to `denoise' the input and reconstruct the uncorrupted
input. The denoising criterion permits overcomplete models (i.e.~models
with more hidden units per layer than input units) to learn useful representations 
by way of predictive opposition between the reconstruction distribution and the regularizer
\citep{vincent2010stacked}.

\subsection{Dropout}
\label{sec:dropout}
Dropout \citep{hinton2012improving,srivastava2014dropout} was motivated
by the idea that sexual reproduction increases the overall fitness of a species
by preventing complex co-adaptations of genes. Likewise, Dropout aims to increase
generalization performance by preventing complex co-adaptations of hidden units.

Formally, consider a feed forward neural network with $L$ hidden layers. 
Define $\h^{(l)}$ as the output vector from hidden layer $l$.
The learning procedure learns a parameterized function $f(\x) = \y$.
Let $\x \in \Reals^{n^x}$ be the input vector, $\h^{(l)} \in \Reals^{n^l}$ be 
the hidden vector, and $\y \in \Reals^{n^y}$ the output vector where
$n^x, n^l, n^y$ are the number of input, hidden (for layer $l$) and output dimensions.
By convention, we define $\h^{(0)} = \x$ and $\h^{(L+1)} = \y$.
Next, define $\W^{(l)} \in \Reals^{n^l \times n^{l-1}}$ 
as the weights and $\bias^{(l)} \in \Reals^{n^l}$ as the
hidden biases for layer $l$. The hidden unit and output activations can be written as
\begin{equation}
  \label{eqn:hid}
  \h^{(l)} = \sigma^{(l)}\left( \W^{(l)} \h^{(l-1)} + \bias^{(l)} \right)
\end{equation}
\begin{equation}
  \label{eqn:ffnet}
  f(\x) = \y = \sigma^{(y)}\left( \W^{(L+1)} \h^{(L)} + \bias^{(L+1)} \right) 
\end{equation}
where $\sigma^{(l)}, \sigma^{(y)}$ are the hidden and output activation functions.
The weight matrices and bias vectors are randomly initialized and learned
via optimization of a cost function, typically via gradient descent.

Under Dropout, multiplicative noise is applied to the hidden activations
during training for each presentation of an input.
This is accomplished by stochastically dropping (or masking) individual hidden
units in each layer \citep{srivastava2014dropout}.
Formally, define $\rand^{(l)} \in \Reals^{n^l}$ such that $r^{(l)}_i
\sim \text{Bernoulli}(p)$,
i.e. $\rand^{(l)}$ is a vector of independent Bernoulli variables where each has probability
$p$ of being 1 \citep{srivastava2014dropout}.
During training, Equation \ref{eqn:hid} becomes
\begin{equation}
  \hat \h^{(l)} = \sigma^{(l)}\left( \W^{(l)} \left(\h^{(l-1)} \circ \rand^{(l-1)}\right) + \bias^{(l)} \right) 
\end{equation}
where $\circ$ denotes the elementwise product.
No masking is performed at test time. In order to compensate for the lack
of masking, the weights are scaled by $p$ once training is complete \citep{srivastava2014dropout}.
Recent results have shown that using Gaussian noise with $\mu = 1$ (as opposed to Bernoulli)
leads to improved test performance \citep{srivastava2014dropout}. At test time each
$r_i = \mu = 1$ and, as such, the weights of the testing model do not require scaling.

The Bernoulli Dropout  procedure can be interpreted as a means of training a different
sub-network for each training example, where each sub-network contains
a subset of the connections in the model. There is an exponential
number of such networks, and many may not be visited during training.
However, the extensive amount of weight sharing between these networks
allows them to make useful predictions regardless of the fact that they
may have not been trained explicitly \citep{hinton2012improving}.
Under this interpretation, the test procedure can be seen as an approximation of the geometric 
average of all the sub-networks \citep{srivastava2014dropout}.

Dropout has inspired a variety of both theoretical and experimental research.
Subsequent theoretical work has found that training with Dropout is equivalent
to an adaptive version of $L_2$ weight decay \citep{wager2013dropout}. Where
the traditional $L_2$ penalty is spherical in weight space, Dropout is akin
to an axis aligned scaling of the $L_2$ penalty such that it takes the curvature
of the likelihood function into account \citep{wager2013dropout}.
An extension of Dropout, DropConnect, applies the mask not to the hidden
units but to the connections between units and was found to
outperform Dropout on certain image recognition tasks \citep{wan2013dropconnect}.
Maxout networks \citep{goodfellow2013maxout} are an attempt to design
a new type of activation function which exploits the benefits of the Dropout
training procedure.

\subsection{Gated Models}
\label{sec:gate}
Where classical neural networks contain only first order interactions among input variables and hidden variables, gated models permit a \emph{tri-partite} graph that connects hidden variables to pairs of input variables. The messages sent in such networks involve multiplicative interactions among variables and permit the learning of structure in the relationship between inputs rather than the structure of inputs themselves \citep{memisevic2007unsupervised,taylor2009factored,memisevic2013learning}. Recent applications of such models include modeling transformations between images \citep{memisevic2013learning},
 3D depth \citep{konda2013learning}, and time-series data
\citep{michalski2014modeling}.

The objective of a typical first order neural network is to
learn a mapping function between an input $\x$ and an output
$\y$. In other words, training involves updating the model
weights such that the model learns the function $f(\x) = \y$.
Instead of learning a mapping between a single input and output
vector, gated models are conditional on a second input $\z$ such that
the learned function is $f(\x|\z) = \y$. We refer to $\z$ as the ``context''.
Instead of a fixed weight matrix as in classical neural networks,
the weights in a gated model can be interpreted as being a
function of the context $\mathbf{z}$  \citep{memisevic2013learning}.

Formally, the hidden activation in a single layered factored gated model with input vectors
$\x \in \Reals^{n^x}$, context $\z \in \Reals^{n^z}$ and factor size $n^f$
\begin{equation}
  \label{eq:fact}
  \h = f(\x | \z) = \sigma^{(h)} \left( \W^T \left( \U \x \circ \V \z \right) + \bias^{(h)} \right)
\end{equation}
where $\U \in \Reals^{n^f \times n^x}$, $\V \in \Reals^{n^f \times n^z}$,
$\W \in \Reals^{n^f \times n^h}$ are learned weight matrices, $\bias^{(h)} \in \Reals^{n^h}$ are the
hidden biases, $\circ$ denotes an elementwise product, and $\sigma^{(h)}$ is the hidden
activation function (typically of the sigmoidal family or piecewise linear).

Note that in the special case where $\z$ is constant, Equation \ref{eq:fact} becomes 
\begin{equation}
  \label{eq:fact_fixed}
  \h = \hat f(\x) = \sigma^{(h)} \left( \hat \W \x  + \bias^{(h)} \right)
\end{equation}
where $\hat \W = \left( \W \circ \left(\left(\V\z\right) \otimes \e\right) \right) \U$, such that 
$\e$ is a $n^h$ dimensional vector of ones
and $\otimes$ is the Kronecker product. Thus, with constant $\z$ the gated model is equivalent to
a feed forward model.

\section{FaMe Training}
The proposed Factored Mean training procedure (FaMe) aims to make use
of the weight modulation property of gated networks as a means
of regularization. The FaMe architecture, like a neural network trained with Dropout,
aims to learn a mapping $f(\x) = \y$ between input $\x$ and output
vector $\y$ (e.g.~for classification). Where Dropout applies
multiplicative noise to the hidden activations during training,
under FaMe training each weight matrix is decomposed into two matrices (or ``factor loadings'') and the
multiplicative noise is applied directly after the input vector is projected onto the first of these matrices.

\begin{figure}
  \begin{center}
    \includegraphics[scale=0.25]{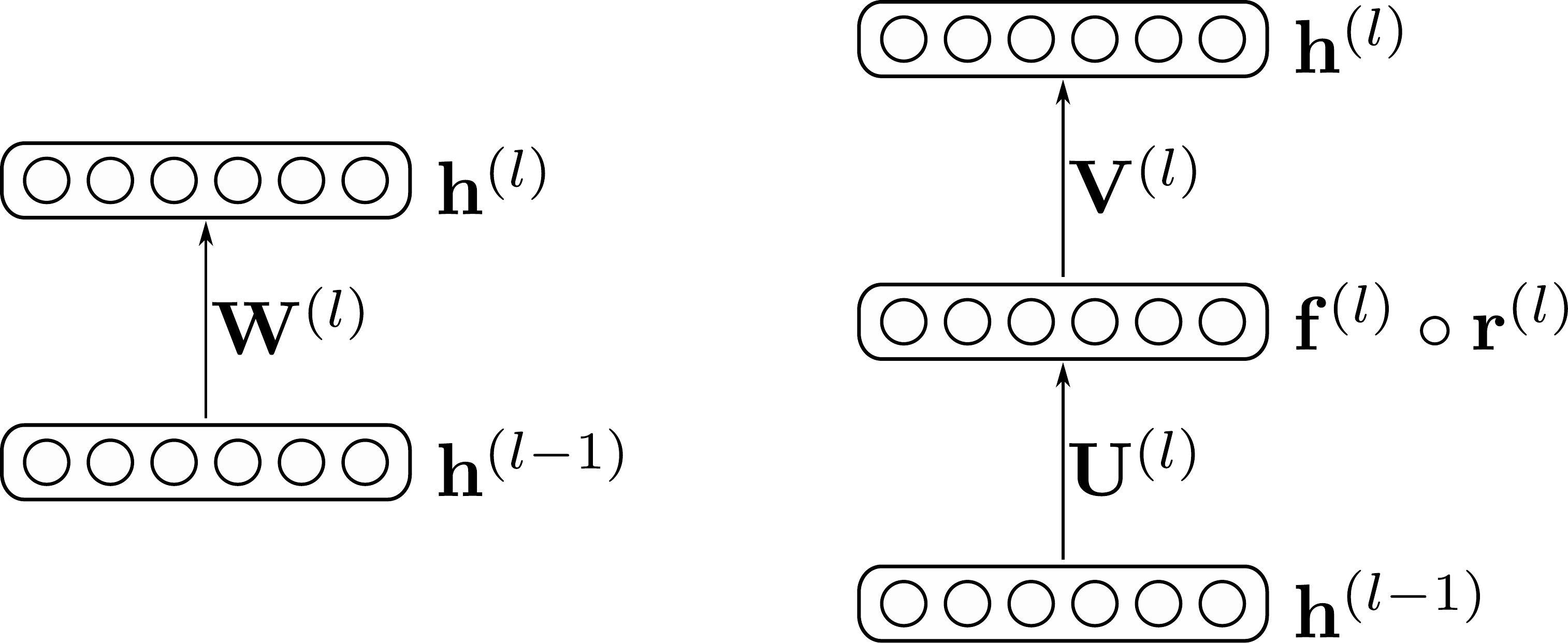}
  \end{center}
  \caption{%\gwt{Can you break this in two so that we can subcaption them?}
  Comparison of a classical feed forward neural network and FaMe.
  The figure on the right depicts the connectivity between hidden layers
  in a typical feed forward neural network. On the left is the connectivity
  between hidden layers in a FaMe model where $\mathbf f^{(l)} = \U^{(l)}\h^{(l-1)}$ 
  and $\rand^{(l)}$ is a vector in independent samples from some noise distribution.
  Biases are omitted for clarity.
  }
  \label{fig:models}
\end{figure}

Formally, given a feed forward neural network with $L$ layers as
described in section \ref{sec:dropout}, instead of a single weight
matrix $\W^{(l)}$ for layer $l$ we define matrices $\U^{(l)} \in \Reals^{f^{l} \times n^{(l-1)}}$,
$\V^{(l)} \in \Reals^{n^l \times f^l}$ where $f^l$ is a free parameter.
Thus, the hidden activation becomes
\begin{equation}
  \label{eqn:mega}
  \h^{(l)} = \sigma^{(l)}\left( \V^{(l)} \left ( \U^{(l)} \h^{(l-1)} \circ \rand^{(l)} \right ) + \bias^{(l)} \right)
\end{equation}
where $\rand^{(l)} \in \Reals^{f^l}$. During training, each $r_i^{(l)}$ is an 
independent sample from a fixed probability distribution (i.e. Bernoulli, Gaussian).
See Figure \ref{fig:models} for a comparison of the FaMe architecture and 
a classical feed forward model.

Note the similarity between Equation \ref{eqn:mega} and the gated model Equation \ref{eq:fact}. Instead
of the model weights being a function of a secondary input vector $\z$, the
FaMe training procedure modulates the weights via a random vector $\rand^{(l)}$.
Like Dropout, the FaMe model can be viewed as training a unique model for
each training example and epoch. In other words, the FaMe model represents a manifold
of models where the settings of $(\rand^{(1)}, \rand^{(2)}, \ldots \rand^{(L+1)})$
can be thought of as the coordinates of a given model on that manifold.

Where Dropout computes an average mean network by scaling the weights by $p$,
at test time we can calculate the mean network learned by FaMe training
by setting each $r^{(l)}_i$ to the expectation of its sampling distribution. This
is similar to Equation \ref{eq:fact_fixed} when we fix secondary input in
the gated model.
However, if we define $r_i^{(l)} \sim N(1, 1)$ (or any other distribution with mean
1), then our hidden activations in our mean network at test time can be simplified to
\begin{equation}
  \h^{(l)} = \sigma^{(l)}\left( \W^{(l)} \h^{(l-1)} + \bias^{(l)} \right)\\
  \label{eqn:FaMetest}
\end{equation}
where $\W^{(l)} = \V^{(l)}\U^{(l)}$ (so, $\W^{(l)} \in \Reals^{n^l \times n^{(l-1)}}$). This
is equivalent to Equation \ref{eqn:hid} of a classical feed forward network.
In effect, the FaMe training procedure
learns a decomposition of the weight matrix $\W^{(l)}$ such that it is robust to
noise. Notice that in the process of learning a decomposition of $\W^{(l)}$, the rank of
$\W^{(l)}$ can be bounded via the choice of $f^l$. In order to not restrict
the rank of $\W^{(l)}$ and not introduce any unnecessary parameters, we
can set $f^l = \min(n^l, n^{(l-1)})$.

The parameters in the FaMe model can be learned using the same
gradient-based optimization
methods as a classical neural network. 

\subsection{FaMe Convolution layers}
A similar technique can be applied to convolution layers, where a single convolution
step can be decomposed into two linear convolution operations. Under FaMe training,
multiplicative noise can be applied after the first convolution. This is similar
to the Network in Network (NIN) model \citep{lin2014network}. Where each filter in NIN can be seen as a small
non-linear MLP, in a FaMe convolution layer the filters are two layer linear networks
with multiplicative noise added after the first layer.

\section{Experiments}

The effectiveness of the FaMe training procedure for image classification 
was evaluated on the MNIST \citep{lecun1998mnist} and CIFAR
datasets \citep{krizhevsky2009learning}. 

\subsection{MNIST}
The MNIST dataset
consists of 60,000 training and 10,000 test examples, each of which
is a 28 $\times$28 greyscale image of a handwritten digit. The training set
was further randomly partitioned into a 50,000 image training set
and a 10,000 image validation set.

Experiments were implemented in Python using the Theano library \citep{Bastien-Theano-2012,bergstra2010theano}.
The model hyper-parameters were chosen based on performance on the validation
set. During training, multiplicative Gaussian noise was a applied
to both the input ($\sim \mathcal{N}(1, 0.5)$) and linear factor
layers ($\sim \mathcal{N}(1,1)$).
Incoming weight vectors for each unit were constrained to a maximum
$L_2$ norm of 2.0. Training was performed using mini-batch gradient descent
on cross-entropy loss with a batch size of 250. Learning rates were annealed by a factor of 0.995
for each epoch. Nesterov accelerated gradient \citep{sutskever2013importance}
was used for optimization, with initial value of 0.5 and increasing
linearly for a set number of epochs. The final momentum value along with
the number of epochs until reaching the final momentum value were chosen based
on validation performance. 

Training was performed for 500,000 weight updates (i.e.~minibatches). The hyper-parameters which
resulted in the lowest validation error were then used to train the final model
on all 60,000 training examples. Again the test model was trained for a total of 500,000 weight updates.

Results on MNIST data are summarized in Table \ref{tab:MNIST}. Using the best
settings of the hyper-parameters found during validation, the test model was 
trained from 10 random initializations and the average error of the ten models
is reported. FaMe training outperforms Dropout \citep{srivastava2014dropout}  and Maxout 
\citep{goodfellow2013maxout}, achieving a best test error of $0.91\% (\pm0.02)$.
We found that restricting the rank of $W$ was beneficial with larger hidden layer sizes,
where the two layer model restricted the size of the linear factor layer to 440.
Additionally, training a model with fewer parameters achieves similar results.
If we consider the 
size of the test model (after the $U$ and $V$ linear projections have been collapsed to
as single weight matrix $W$) a FaMe model
with 3 hidden layers with 512 units per layer has $\sim$93K effective parameters.
The maxout model
has $\sim$2.3 million free parameters and Gaussian Dropout has $\sim$1.9 million parameters.

\begin{table}
  \centering
  \begin{tabular}{cccc}
    \hline
    \textbf{Method} & \textbf{Unit Type} & \textbf{Architecture} & \textbf{Test Error \%}\\
    \hline
    Bernoulli Dropout NN + weight constraints & ReLU & 2 layers, 8192 units & 0.95\\
    (Srivistava et al., 2014)\\
    Bernoulli Dropout NN + weight constraints & Maxout & 2 layers, (5x240) units & 0.94\\
    (Goodfellow et al., 2013)\\
    Gaussian Dropout NN + weight constraints & ReLU & 2 layers, 1024 units & 0.95 +/- 0.05\\
    (Srivistava et al., 2014)\\
    \hline
    Gaussian FaMe NN + weight constraints & ReLU & 2 layers, 1024 units & \textbf{0.91 +/- 0.02}\\
    Gaussian FaMe NN + weight constraints & ReLU & 3 layers, 512 units & 0.92 +/- 0.04\\
  \end{tabular}
  \caption{Result of MNIST classification task. FaMe training outperforms dropout, even
  when training models with far fewer parameters.}
  \label{tab:MNIST}
\end{table}

\subsection{CIFAR}
The CIFAR datasets \citep{krizhevsky2009learning} contain 60,000 $32\times32$ color images, 50,000
of which are training examples with the remaining 10,000 used for testing.
The CIFAR-10 dataset contains images from 10 classes where the CIFAR-100 contains
100 classes.
Hyper-parameter selection followed a similar procedure to that used above for MNIST
classification, further partitioning the training set into 40,000 training and 10,000 validation
images. Our CIFAR models consist of three convolutional FaMe layers
followed by two fully connected FaMe layers. The results are summarized in Table \ref{tab:CIFAR10}.
For image preprocessing, we follow the same procedure as \citet{srivastava2014dropout} and \citet{goodfellow2013maxout}.
Global contrast normalization over each color channel is followed by ZCA whitening.

Although FaMe fails to outperform Dropout or Maxout, it is
competitive with other recent results on CIFAR-10 and CIFAR-100.
The discrete subset of parameters we considered for hyper-parameter
cross-validation was substantially smaller
than those considered for the MNIST task.
We expect that with more time to optimize the parameters, potentially using
tools such as Bayesian hyper-parameter optimization
\citep{snoek2012practical}, performance will improve.
\begin{table}
  \centering
  \begin{tabular}{lcc}
    \hline
    \textbf{Method} & \textbf{CIFAR-10 Error \%} & \textbf{CIFAR-100 Error \%}\\
    \hline
    Conv Net + Spearmint \citep{snoek2012practical} & 14.98 & - \\
    Conv Net + Dropout (fully connected) \citep{srivastava2014dropout} & 14.32 & 41.26\\
    Conv Net + Dropout (all layers) \citep{srivastava2014dropout} & 12.61 & 37.20 \\
    Conv Net + Maxout  \citep{goodfellow2013maxout} & 11.68 & 38.57\\
    Conv Net + FaMe & 12.85 & 39.79\\
  \end{tabular}
  \caption{Comparison of FaMe training with other models on the CIFAR-10 dataset}
  \label{tab:CIFAR10}
\end{table}

\section{Comparison with Dropout training}
We conducted a separate set of experiments on MNIST to further probe
the learning dynamics of FaMe vs.~Dropout.  All experiments described in this section were
performed using a model with two non-linear (ReLU) hidden layers and
multiplicative Gaussian noise from $\mathcal{N}(1,0.5)$ on the input
and $\mathcal{N}(1,1)$ on the hiddens or factors.  For both FaMe and
Dropout training, hyper-parameters were chosen based on validation set
performance.

\subsection{Avoiding Overfitting}
\begin{figure}
  \begin{center}
    \includegraphics[scale=0.4]{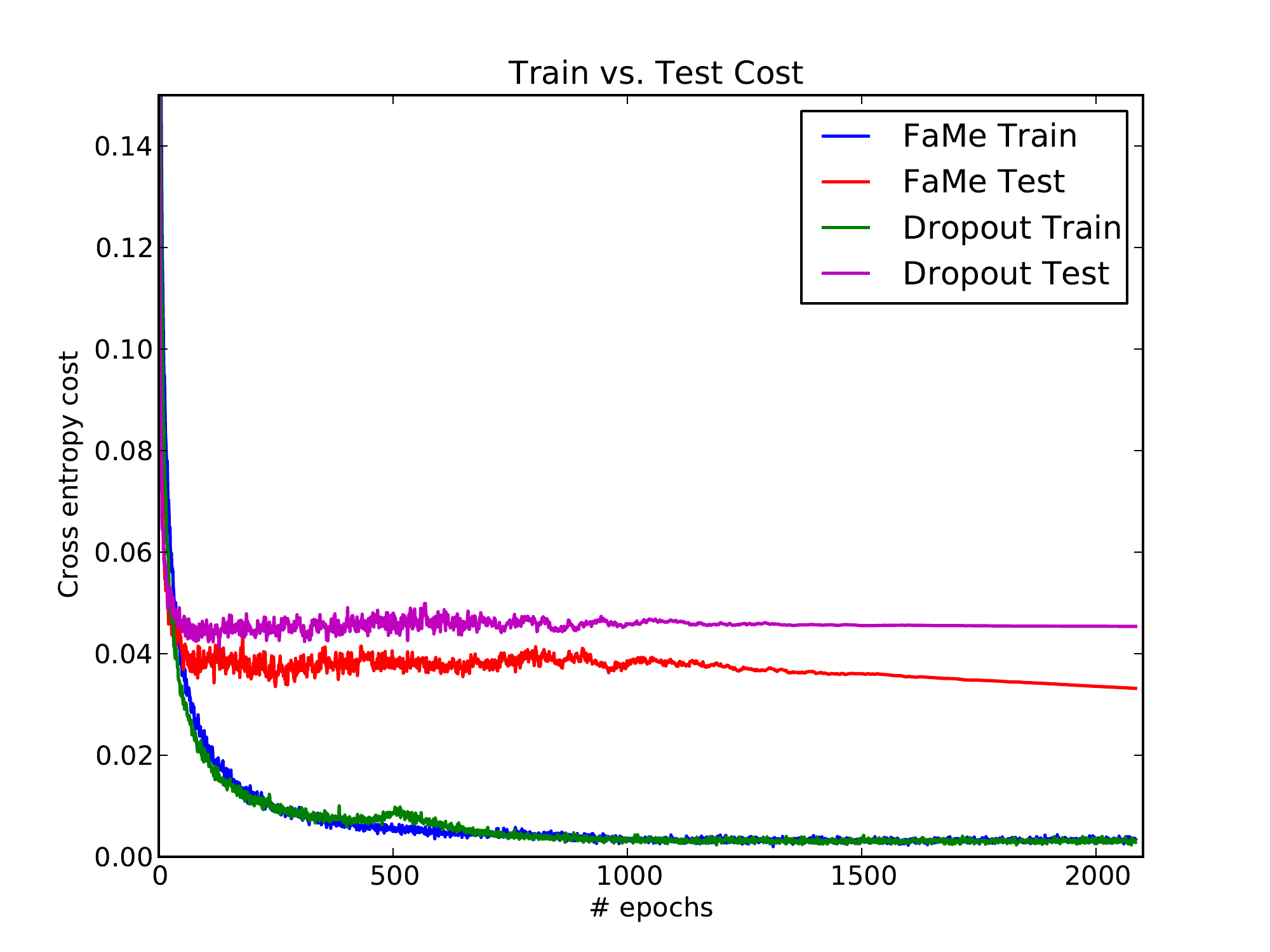}
  \end{center}
  \caption{Training cost vs. test cost as training progresses on
  an example training run for both FaMe and Dropout.
  Notice that with FaMe training, the
  test cost continues to decrease for the entire duration of training.}
  \label{fig:training}
\end{figure}
Similar to Dropout training \citep{srivastava2014dropout},
FaMe training prevents overfitting and, as such, does not require
early stopping. As seen in Figure \ref{fig:training} the test
cost continues to decrease for the entirety of training.
Dropout is also successful at avoiding overfitting, i.e.~the test
cost does not increase with continued training. However, the
test cost plateaus earlier than with FaMe training.

\subsection{Mean testing procedure}
A full mathematical analysis of the mean test procedure given in Equation \ref{eqn:FaMetest}
is difficult due to the non-linearity of the hidden units. As such, we experimentally
verify that the testing procedure is indeed approximating a geometric mean of 
predictions of the noisy subnetworks visited during training.
Using the test data as input, we generate samples output from these
noisy networks. The mean prediction can be calculated
by first computing the geometric mean of the sample outputs and comparing
to the output given by our deterministic testing procedure.

\begin{figure}
  \begin{center}
    \includegraphics[scale=0.35]{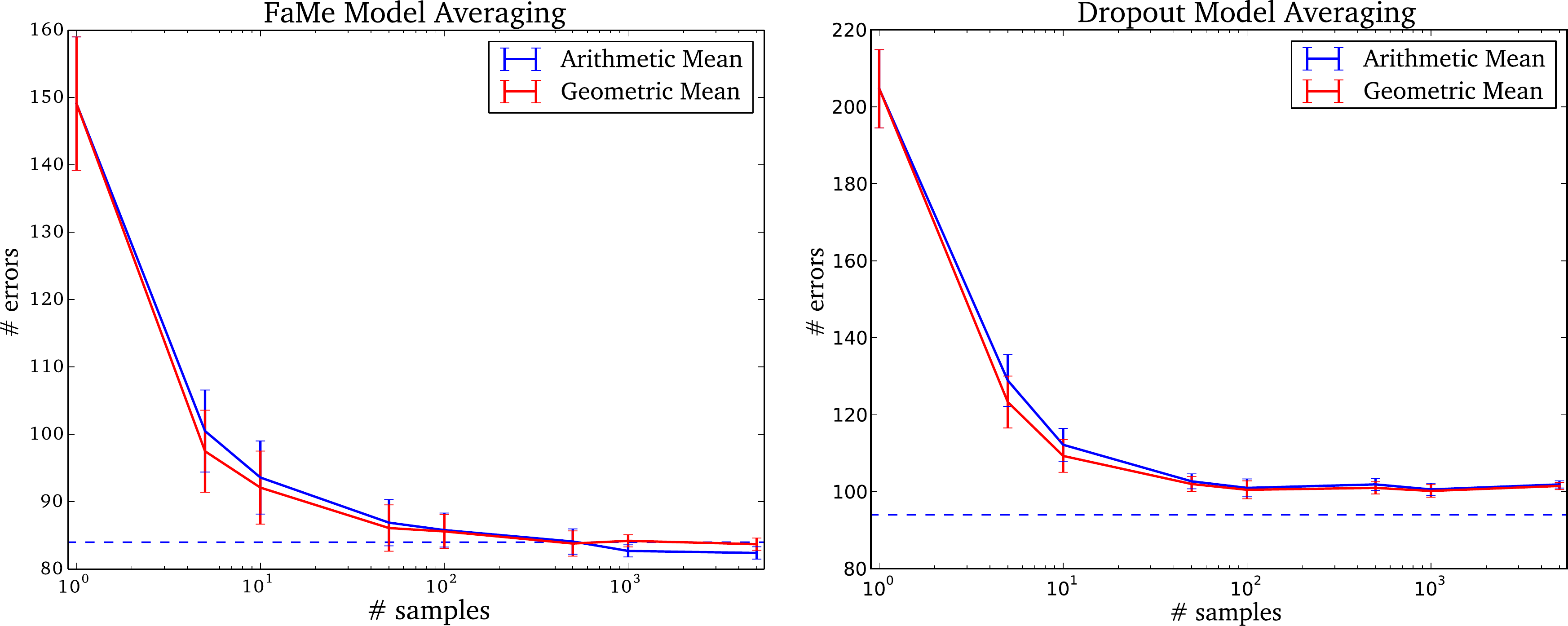}
  \end{center}
  \caption{Comparison of the FaMe and Dropout test procedure
  to the true arithmetic and geometric means as estimated
  via sampling outputs of random subnetworks.
  With FaMe training (left), the prediction of the testing procedure (dashed blue line) does in fact give a good estimate
  of the true geometric mean prediction.
  }
  \label{fig:test_proc}
\end{figure}

Figure \ref{fig:test_proc} demonstrates that the FaMe testing procedure
gives an accurate estimate of the true geometric mean of all subnetworks.
By comparison, even though the Dropout test procedure outperforms the estimated 
geometric mean of the subnetworks, it does not provide a good estimate
of the true mean network. 

Note that both the FaMe and Dropout models contained
two non-linear (ReLU) hidden layers and were trained with multiplicative
Gaussian noise.
Interestingly, the arithmetic mean prediction of the FaMe subnetworks appears
to slightly outperform geometric
mean prediction which may warrant further investigation.

\subsection{Adaptive $L_2$ weight decay}
\begin{figure}
  \begin{center}
    \includegraphics[scale=0.42]{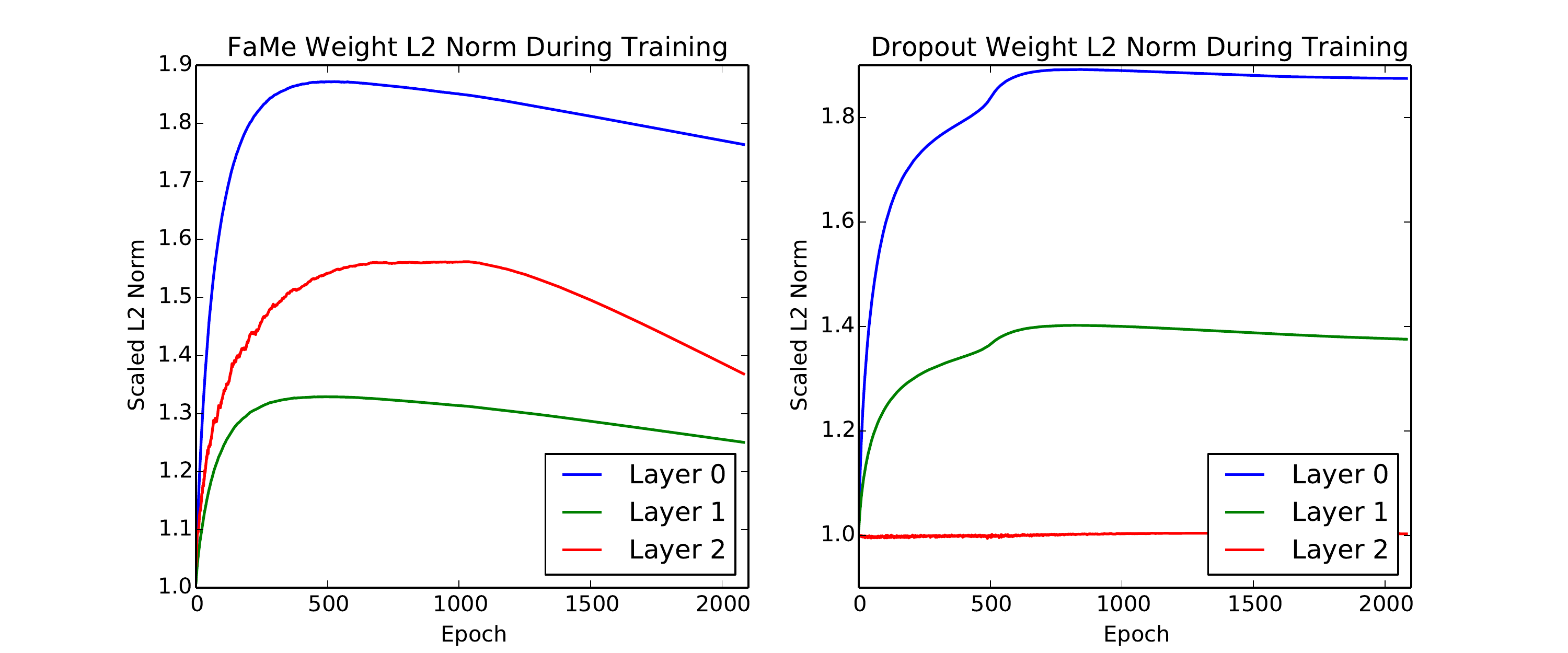}
  \end{center}
  \caption{$L_2$ Norm of weight matrices during training. For visualization
  the $L_2$ norms are scaled by their value at initialization.
  Both training procedures exhibit behavior similar to $L_2$ weight decay,
  but the result is more pronounced in the case of FaMe training.
  }
  \label{fig:l2}
\end{figure}

% Although the equivalence between small amounts of additive Gaussian
% noise and $L_2$ weight decay is well understood \citep{bishop1995training},
% the link between $L_2$ weight decay and Dropout was not made until
% about a year after its introduction.
\citet{wager2013dropout} have 
shown Dropout to be equivalent to adaptive $L_2$ weight decay.
To examine the link between weight decay and both FaMe and Dropout training,
we monitor the $L_2$ norms of the weight matrices of under
both training regimes. Figure \ref{fig:l2} depicts the evolution
of the $L_2$ norm of each $W$ weight matrix during training, relative
to $L_2$ norm of its initial value (i.e.~before training).
Note that under FaMe trianing, we plot the $L_2$ norm of the implied
weight matrix $W = VU$, where $V$ and $U$ are the factored weights
that are actually learned by the model. Although no explicit $L_2$ penalty
was used, both FaMe and Dropout training seem to impose an implicit
penalty on the $L_2$ norm of the weight matrices. However,
the effect is more pronounced under FaMe training.

\section{Conclusion}
Regularization is essential when training large neural networks. Their
power as universal function approximators allows them to memorize sampling
noise in the training data, leading to poor generalization to unseen data.
Traditional forms of generalization are means of limiting the model's capacity.
These include early stopping, weight decay, weight constraints, and addition
of noise. Currently, Dropout (and related methods such as DropConnect) are
the most effective means of regularizing large neural networks. These
amount to efficiently visiting a large number of related models at training time, while
aggregating them to a single mean model at test time. The proposed
FaMe model aims to apply this same reasoning, but can be interpreted as a special form of weight-modulating, gated architecture.

Like Dropout, FaMe visits a family of models during training while allowing
an efficient testing procedure for making predictions.
Models trained with FaMe outperform Dropout training, even when such models
have an order of magnitude fewer effective parameters. Additionally, restricting the rank of the factor loadings
can be used as a means of controlling the number of free parameters.
This is supported by recent work has looked closely at the significant redundancy in the parameterization
of deep learning architectures, and proposed low-rank weight matrices as a way to massively reduce parameters while preserving predictive accuracy \citep{denil2013predicting}.

\bibliography{citations}{}
\bibliographystyle{iclr2015}

\end{document}